\documentclass[runningheads]{llncs}

\usepackage{eccv}

\usepackage{eccvabbrv}

\usepackage{graphicx}
\usepackage{booktabs}
\usepackage{mathbbol}
\usepackage{multirow}
\usepackage{algorithm}
\usepackage{algorithmic}
\usepackage[normalem]{ulem}
\useunder{\uline}{\ul}{}
\usepackage[accsupp]{axessibility}  

\usepackage{hyperref}

\usepackage{orcidlink}

\begin{document}

\title{CLIP is Also a Good Teacher: A New Training Framework for Inductive Zero-shot Semantic Segmentation} 

\titlerunning{Abbreviated paper title}

\author{Jialei Chen\inst{1}\and
Daisuke Deguchi\inst{1}\and
Chenkai Zhang\inst{1}\and Xu Zheng\inst{2}\and Hiroshi Murase\inst{1}}

\authorrunning{F.~Author et al.}

\institute{Nagoya University \and
HKUST}

\maketitle

\begin{abstract}
  Generalized Zero-shot Semantic Segmentation aims to segment both seen and unseen categories only under the supervision of the seen ones. To tackle this, existing methods adopt the large-scale Vision Language Models (VLMs) which obtain outstanding zero-shot performance. However, as the VLMs are designed for classification tasks, directly adapting the VLMs may lead to sub-optimal performance. Consequently, we propose \textbf{CLIP-ZSS} (Zero-shot Semantic Segmentation), a simple but effective training framework that enables \textbf{any image encoder} designed for closed-set segmentation applied in zero-shot and open-vocabulary tasks in testing \textbf{without combining with VLMs or inserting new modules.} CLIP-ZSS consists of two key modules: Global Learning Module (GLM) and Pixel Learning Module (PLM). GLM is proposed to probe the knowledge from the CLIP visual encoder by pulling the CLS token and the dense features from the image encoder of the same image and pushing others apart. Moreover, to enhance the ability to discriminate unseen categories, PLM consisting of pseudo labels and weight generation is designed. To generate semantically discriminated pseudo labels, a multi-scale K-Means with mask fusion working on the dense tokens is proposed. In pseudo weight generation, a synthesizer generating pseudo semantic features for the unannotated area is introduced. Experiments on three benchmarks show large performance gains compared with SOTA methods. 
\end{abstract}

\section{Introduction}
Semantic segmentation stands as a cornerstone task in the field of computer vision and has witnessed significant strides in performance and capability, thanks to the swift advancements in deep learning methodologies~\cite{resnet,vggnet,transformer}. 
Inherently, to learn high-performance neural networks for semantic segmentation, the requirements of an extensive dataset with precise annotations are imperative.
However, collecting such data is extremely expensive and time-consuming, \eg, each image in the Cityscapes \cite{cityscapes} dataset takes 1.5 hours to annotate accounting for quality control. 
Accordingly, researchers have focused on applying Vision-Language Models (VLMs) like CLIP~\cite{clip}, ALIGN~\cite{ALIGN}, and RAM~\cite{recognizeanything} for zero-shot tasks to obviate the need for high-quality annotations.

\begin{figure}[tb]
\centering
\begin{minipage}[h]{0.49\linewidth}
\centering
\includegraphics[width=1\linewidth]{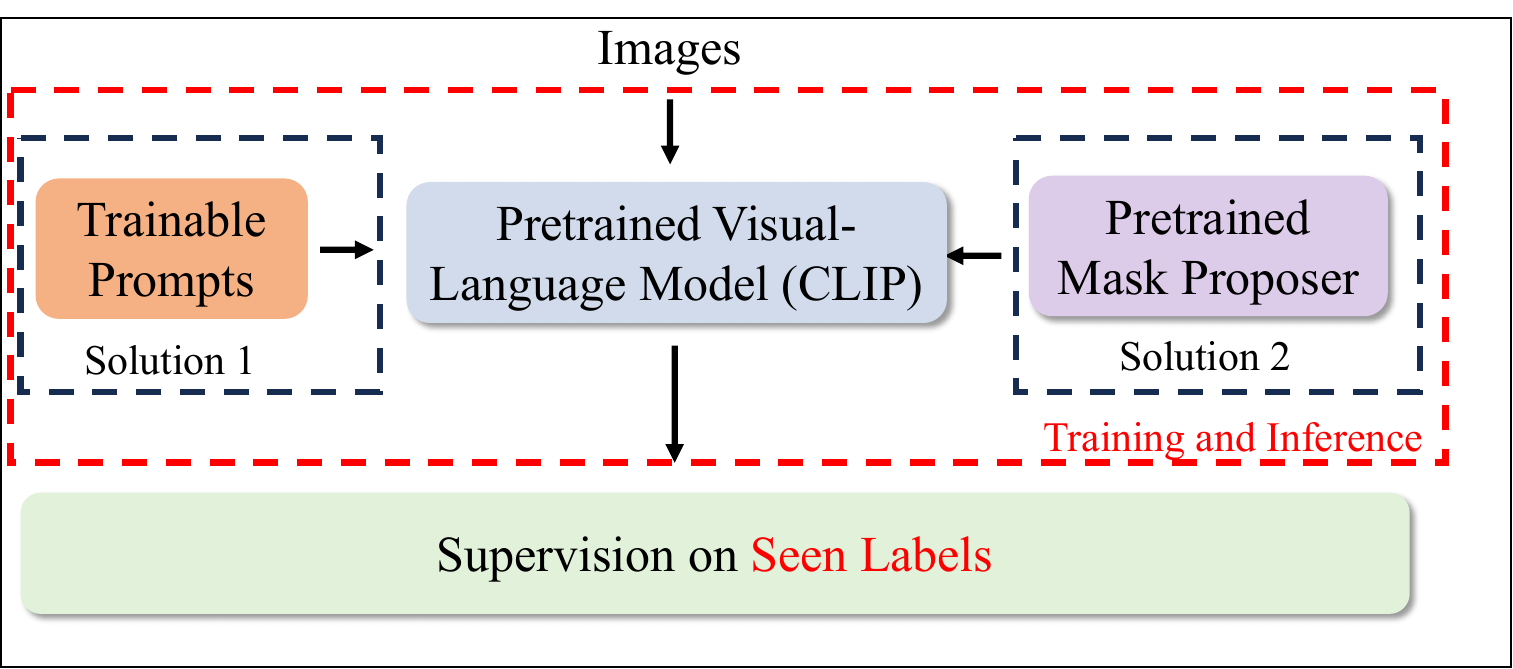}
\subcaption{Adopting by \textcolor{orange}{prompts} or \color{violet}{mask proposer}}
\label{problema}
\end{minipage}
\begin{minipage}[h]{0.49\linewidth}
\centering
\includegraphics[width=1\linewidth]{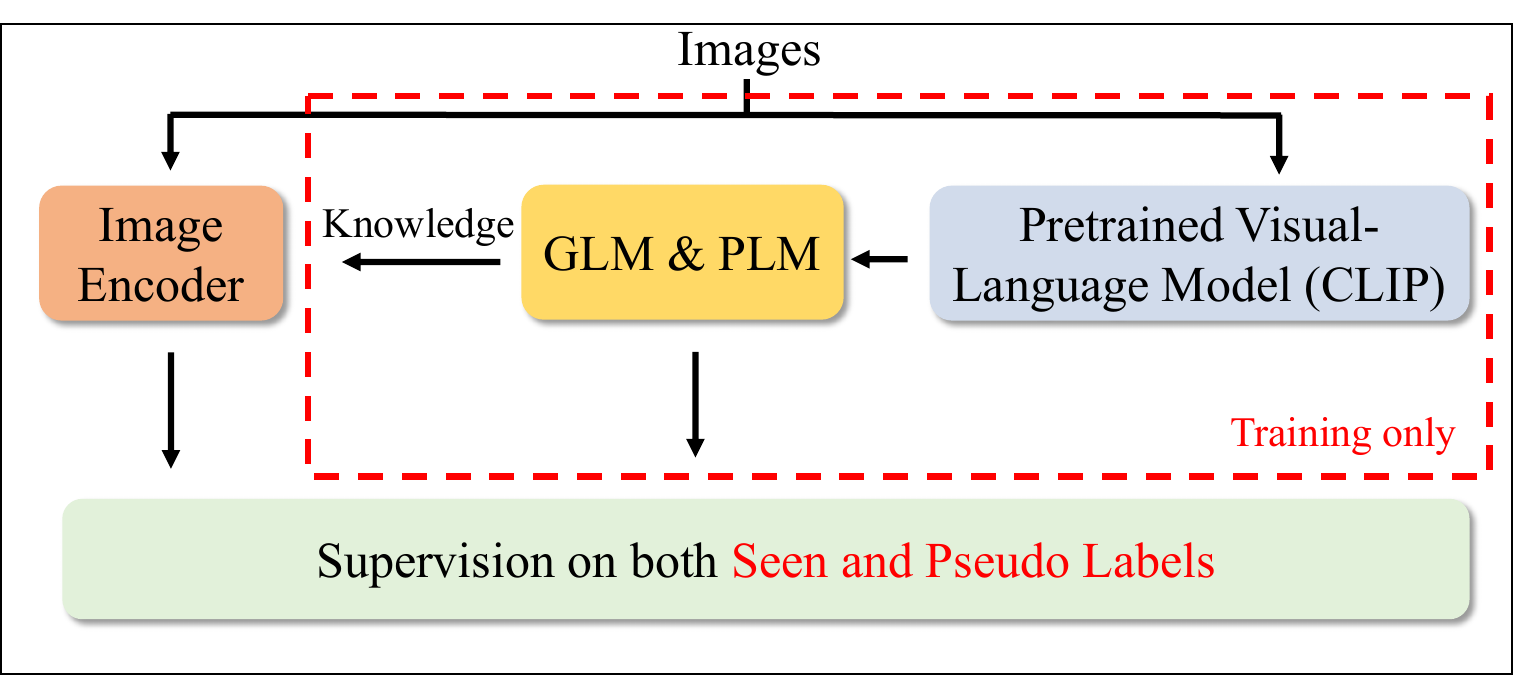}
\subcaption{Our methods}
\label{problemc}
\end{minipage}

\caption{(a) Existing adopting-based GZLSS methods introduce new modules: \textcolor{orange}{trainable prompts} or {\color{violet}{mask proposer}} to adapt the VLMs for segmentation. (b) Our methods transfer VLMs' knowledge to image encoders for close-set segmentation while introducing no new modules to the image encoder.}
\label{problems}
\vspace{-15pt}
\end{figure}

Though remarkable, these models cannot localize and classify multiple objects in images, indicating the ineffectiveness for downstream dense prediction tasks, \ie, semantic segmentation. To mitigate this issue, some early works \cite{Lseg,Openseg} try to fine-tune the pre-trained VLMs directly. Recently, inspired by the Visual Prompt Tuning (VPT) \cite{visualprompttuning}, and the appearance of the adapter structure \cite{adapter}, as shown in Fig.~\ref{problema}, some work that effectively takes advantage of two strategies also attracted much attention \cite{zegclip,sideadapter,maskclip,mvpseg,embeddingmodulation}. Another solution is based on the pre-trained \cite{simplebaseline,namedmask,masqclip} or online learning mask proposer \cite{freeseg,panopticmaskclip,zegformer} to train a new segmentation model or fine-tune the VLMs in the object-level. Despite the pioneer researchers' huge steps, prevalent works still suffer from the following drawbacks: \textbf{1)} For the adapter-based methods, the VLMs are designed for classification tasks, leading to sub-optimal performance for downstream tasks. \textbf{2)} For the mask proposer-based methods, the training process is split into several steps, \ie, proposer training, and VLM adopting, and the quality of proposals highly affects the performance. \textbf{3)} There is a great waste of information in the unannotated areas, which constrains the model's performance.

To address these challenges, we propose a simple but effective training framework called \textbf{CLIP-ZSS} (Zero-shot Semantic Segmentation). CLIP-ZSS transfers (teaches) the knowledge of CLIP \textbf{only once} to \textbf{any image encoder designed for closed-set semantic segmentation} to function \textbf{zero-shot and open-vocabulary} tasks without additional modules or combining with VLMs in inference while effectively \textbf{utilizing the unannotated areas}. Except for the \textbf{fixed} CLIP and segmentation image encoders, we propose two modules in CLIP-ZSS: Global Learning Modules (GLM) and Pixel Learning Modules (PLM). 

Besides aligning the dense features from the image encoder with the text features from the CLIP text encoder, GLM pulls \textbf{all} the dense features and the CLIP visual CLS tokens that contain \textbf{global information} of an image together and pushes other CLS tokens from different images apart. 
Though GLM can bridge the gap between vision and text, it fails to offer reliable fine-grained supervision for semantic segmentation. Consequently, we propose the PLM comprised of two novel components: pseudo labels generation and pseudo weights generation. To generate pseudo labels, we approach multi-scale KMeans with mask fusion. Formally, we slide different sizes of windows to average dense tokens from the CLIP visual encoder serving as the mask seeds. Then, a mask fusion algorithm is introduced to fuse the masks corresponding to different mask seeds by their cosine similarities. Note that even without any category information, the fused labels are still \textbf{semantically discriminative}, \ie, masks of different seeds belong to different categories. Despite discriminative, it remains unclear how many unknown categories exist within them and which categories these pseudo-labels belong to. Therefore, a transformer decoder-based pseudo-weight synthesizer is proposed to produce prototypes for these pseudo labels. The dense features are mapped to the semantic space by both the seen text features from the CLIP text encoder and the produced prototypes to achieve the segmentation task.

Our main contributions can be listed as 1) We propose GLM, a simple but effective module to transfer the knowledge of CLIP to any image encoder designed for close-set semantic segmentation. 2) We propose PLM, a novel module to generate semantically discriminative pseudo labels for the unannotated areas and pseudo prototypes to provide additional supervision. 3) We conducted extensive experiments on three benchmark datasets on zero-shot semantic segmentation: PASCAL VOC, COCO-Stuff, and PASCAL Context. Compared with SOTA, our method outperforms 2.2\%, 1.3\%, and 8.5\% in hIoU. Surprisingly, without the help of CLIP, our methods can be adapted to open-vocabulary tasks and achieve impressive performance while \textbf{7 times faster} than SOTA methods at most.

\section{Related Works}
\subsection{Close-set Semantic Segmentation}
Prevalent semantic segmentation can be grouped into two aspects: pixel-level classification and mask-level classification. For pixel-level classification, FCN \cite{FCN} as the first fully convolutional network to employ end-to-end semantic segmentation determines the paradigm of pixel-level semantic segmentation. Since FCN, many works, \eg, DeepLab series \cite{deeplabev3,deeplab}, Deformable convolution \cite{deformableconv}, aim to enlarge the receptive field to further improve the performance of pixel-level methods. With the appearance of self-attention \cite{transformer} and ViT \cite{vit}, many works \cite{setr,segformer,segnext,swin} replace the conventional convolutional backbone to the self-attention-based one and achieved remarkable performance. Another aspect treats the semantic segmentation tasks as a mask classification task. MaskFormer \cite{maskformer} and Mask2Former \cite{mask2former} are two representative works. Specifically, these models first generate several object queries corresponding to objects. Then these object queries are decoupled to do classification and mask prediction tasks, respectively.

However, these methods are designed for close datasets where only predefined categories can be distinguished. Additionally, achieving state-of-the-art performance requires a large amount of annotated high-quality images, which are expensive and time-consuming to collect for training a segmentation model.
\textit{In contrast, our proposed CLIP-ZSS frees the closed-set segmentation model from the limitations of a specified dataset and the need for high-quality data that is expensive to collect while maintaining its segmentation ability.}

\subsection{Zero-shot Semantic Segmentation}
Before the VLMs, \eg, CLIP \cite{clip}, several works tried to bridge the gap between vision and language by projecting the features from vision models to the semantic space which is spanned by the large scale of texts \cite{cagnet,spnet}. With the appearance of large-scale VLMs, represented by CLIP \cite{clip} and ALIGN \cite{ALIGN}, zero-shot vision tasks have entered a new era. Due to the impressive zero-shot ability, researchers aim to transfer this ability to downstream tasks. 

Benefiting from visual prompt tuning \cite{visualprompttuning} and adapter \cite{adapter}, some work introduces additional trainable parameters for the VLMs to adopt the dense tokens to do the downstream tasks, \eg, semantic segmentation \cite{sideadapter,Lseg,Openseg,zegclip,maskclip,mvpseg}. Different from introducing new modules, others finetune the VLMs by introducing explicit mask proposer \cite{freeseg,simplebaseline,namedmask,panopticmaskclip,zegformer} motivated by the success of mask-level semantic segmentation \cite{mask2former,maskformer}. However, both types of finetuning-based methods suffer from the shortage that the VLMs are designed for classification tasks and lead to sub-optimal performance to segmentation. Besides, there is a great waste of the unannotated areas. Motivated by recent advances in understanding the CLIP\cite{PACL} and unsupervised learning \cite{SETGO}, we propose a simple but effective method to transfer the knowledge of CLIP to any pixel-level semantic segmentation models \textbf{without} introducing \textbf{explicit} mask proposer or changing anything. Meanwhile, we can also take advantage of the unannotated areas by generating pseudo-semantic-discriminative labels and pseudo weights. 

The most related works are MaskCLIP \cite{maskclip} and ZS3/ZS5 \cite{zs3}. Different from MaskCLIP, our methods do not rely on CLIP in testing, and our methods are applied to inductive rather than transductive settings. For ZS3/ZS5, they need additional fine-tuning after the training of the image encoder, however, our methods need no fine-tuning after training. Another paper \cite{clipissegmenter}, though generating pseudo labels, has to rely on CAM \cite{cam} and other delicately designed modules proposed in this paper. Our methods do not change or add anything to the original CLIP. Compared with pseudo feature generation methods from semantic to vision as input for finetuning the classifier \cite{zs3,pading}, our pseudo weights are produced from vision to semantic and as classifiers for training the whole model. Moreover, our methods are trained only once rather than multi-steps \cite{zs3,pading}. Meanwhile our pseudo labels are different from those provided by well-known self-training algorithms \cite{zs3,zegclip,maskclip} that annotate with trained models by the unseen category names nor the ones which are generated based on multi-modality \cite{pseudoboundingboxlabels,crossmodallabeling}.

\section{Methods}
\subsection{Preliminary.}
\noindent \textbf{Inductive GZSS. } Normally, GZSS applies the images without any unseen category. However, due to the frequent co-occurrence of unseen and seen categories, such settings may not support the model training. Consequently, most researchers refer to Generalized Zero-shot Semantic Segmentation (GZSS) within the setting of Generalized Zero-Label Semantic Segmentation (GZLSS). For inductive settings, the categories in training datasets are separated into two groups: seen categories $\mathcal{A}^s$ and unseen categories $\mathcal{A}^u$. Moreover, there is no overlap between the two groups $\mathcal{A}^s \cap \mathcal{A}^u = \varnothing$. In training, only the pixel-level annotations of $\mathcal{A}^s$ can be accessed and the existence, \ie, how many and what, of $\mathcal{A}^s$ is known. The annotations belonging to the unseen categories are replaced by the same `ignored'. In inference, both the $\mathcal{A}^u$ and $\mathcal{A}^s$ need to be segmented. 

\begin{figure}[tb!]
\centering
\includegraphics[width=\linewidth]{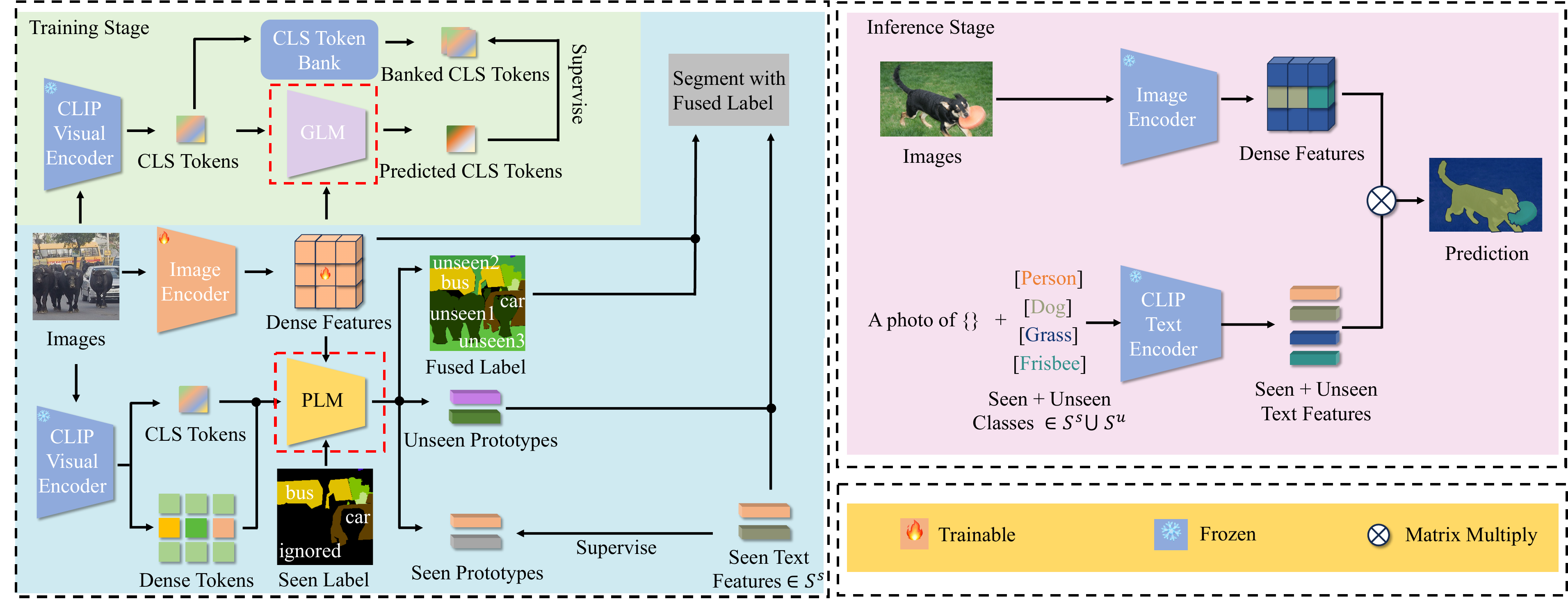}
\caption{The structure of CLIP-ZSS. Proposed methods are indicated by red dot lines.}
\label{methodoverview}
\vspace{-10pt}
\end{figure}

\noindent \textbf{Method overview.} We approach CLIP-ZSS to enable \textbf{any image encoder for close-set segmentation to employ zero-shot and open-vocabulary tasks without introducing any new modules to the image encoder or combining with VLMs}, as shown in Fig. \ref{methodoverview}. In training, besides the frozen CLIP model and an image encoder to extract features, CLIP-ZSS introduces two novel modules termed Global Learning Module (GLM) in Sec. \ref{GLM} and Pixel Learning Module (PLM) in Sec. \ref{PLM}. The GLM aims to directly transfer the knowledge of the CLIP visual encoder to the image encoder to bridge the gap between the vision and semantics through the CLS visual token. However, relying only on GLM can not handle segmentation tasks due to the lack of pixel-level supervision. Consequently, PLM is proposed to generate pixel-level labels which are fused by the seen label and the pseudo labels for unannotated areas. Note even without any category information, the generated pseudo labels are semantically discriminative, \ie, the masks belong to different seeds are different categories. Meanwhile, as the uncertainty of how many and what unseen categories may appear in the dataset, PLM also produces unseen prototypes acting as the classifier to further distinguish between seen and unseen categories. In inference, only the image encoder is applied where images are fed into the image encoder to obtain the dense features and matrix multiplication is done between the dense features and the text features from the CLIP text encoder without any combination with VLMs or inserting new modules.


\subsection{Global Learning Module} \label{GLM}
CLIP \cite{clip}, as one of the most famous Vision-Language Models (VLMs), is trained by aligning the CLS token corresponding to the whole images and ground truth text pairs, which indicates that only the CLS token plays the role of bridging the gap between vision and text. Therefore, to transfer the knowledge into image encoders \textbf{designed for segmentation instead of classification}, we design the Global Learning Module (GLM) inspired by the self-attention structure \cite{transformer,non-local}.

The core design of GLM aims to align between the dense features from the image encoder and the CLS token from the CLIP visual encoder. Specifically, an input image is fed into a trainable image encoder and a frozen CLIP visual encoder separately to obtain the dense features $\mathcal{R}^{B\times C\times L}$ and the CLS token $\mathcal{S}^{B\times C}$ where $B$ indicates the batch size, $C$ is the channel number of output representations and $L=H\times W$ is the multiplication of the height and width of dense features. Different from all the existing self-attention-based methods \cite{transformer,non-local,vit}, we do not introduce any trainable parameters to produce $\mathcal{Q}$, $\mathcal{K}$, and $\mathcal{V}$. The CLS tokens act as the $\mathcal{Q}$ directly and dense features as $\mathcal{K}$ and $\mathcal{V}$. The weight $\mathcal{W}$ is generated by the batch matrix multiply:
\begin{equation}
  \mathcal{W} = \text{Softmax}(\frac{\mathcal{S} * \mathcal{R}}{\sqrt{C}}), 
  \label{weight}
\end{equation}
where $\mathcal{W}$$^{B \times L} \in [0,1]$. `$*$' indicates the batch matrix multiply. The weight $w_i \in \mathcal{W}$ for the $i$th feature $r_i \in \mathcal{R}$ is the softmax along the $L$. Once the weight is generated, we apply the batch matrix multiply to the $\mathcal{W}$ and the $\mathcal{R}$ to produce the predicted CLS token $\hat{\mathcal{S}}$:
\begin{equation}
  \hat{\mathcal{S}} = \mathcal{W} * \mathcal{R}^T. 
  \label{predictedclstoken}
\end{equation}
Motivated by the success of contrastive learning \cite{moco,mocov2,simclr,simclrv2}, we apply the InfoNCE \cite{cpc} as the loss function to do contrastive learning between $\mathcal{S}$ and $\hat{\mathcal{S}}$,
\begin{equation}
  \mathcal{L}_{global} = \Sigma_i^B{\frac{\exp(s^T_i \hat{s}_i / \tau)}{\Sigma_{j \neq i}^{B}{\exp(s_{j}^T  \hat{s}_i) / \tau})+\exp(s^T_i  \hat{s}_i / \tau)}},
  \label{tokendistillation}
\end{equation}
where $\tau$ indicates the temperature for the contrastive loss. $s_i \in \mathcal{S}$ and $\hat{s_i} \in \hat{\mathcal{S}}$ depict the CLS token and reconstructed CLS token of the $i$th image in a batch.

To better learn the knowledge embedded in the CLS token, inspired by the success of memory bank \cite{moco,memorybank}, we propose a CLS token bank $\mathcal{V}$ to store the previous CLS tokens to replace $\mathcal{S}$ to achieve higher performance. Let $\mathcal{V}_n = \left\{\mathcal{S}_{t-1},\mathcal{S}_{t-2}......\mathcal{S}_{t-n}\right\}$ be the CLS token bank where $t$ indicates the current iteration, and $n$ indicates the size of the CLS token bank. During training, before the parameter update, we dequeue the $\mathcal{S}_{t-n}$ in $\mathcal{V}$ and enqueue the $\mathcal{S}_{t}$ to $\mathcal{V}$.

\noindent \textbf{Discussion about GLM. }The core idea for the GZLSS is to recognize the unseen categories without any supervision. To achieve this, most researchers utilize the text features from the CLIP text encoder as the classifier and finetune the \textbf{dense tokens} from the visual encoder. However, CLIP is designed for classification and the dense tokens are not aligned with text features before, which may result in sub-optimal performance in segmentation. GLM transfers the knowledge through \textbf{CLS tokens} from multiple images to \textbf{segmentation models} to solve the issues above. More results about GLM are shown in Section \ref{ablation section}.

\begin{figure*}[t!]
\centering
\begin{minipage}[h]{0.51\linewidth}
\centering
\includegraphics[width=\linewidth]{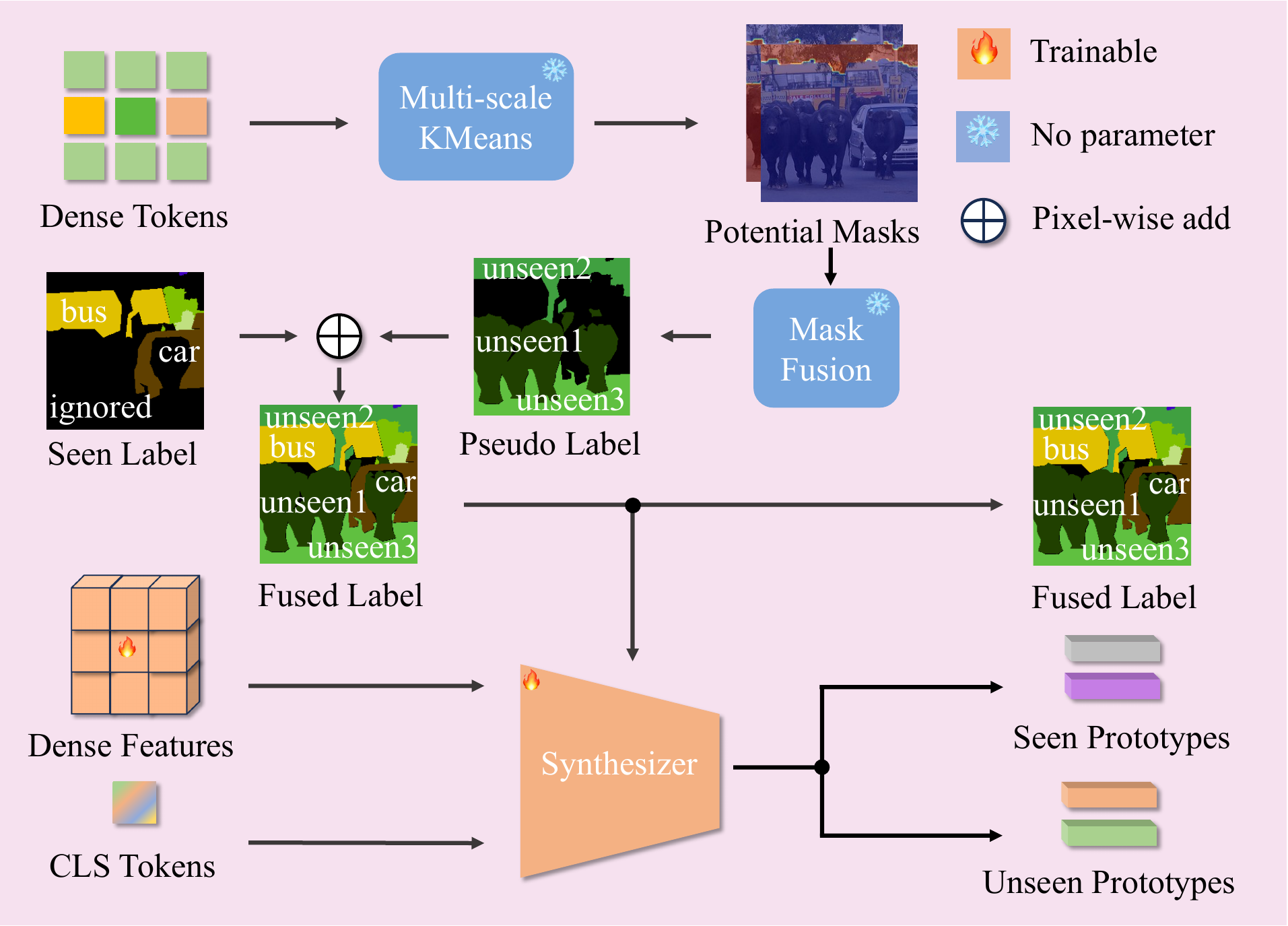}
\subcaption{The overview of Pixel Learning Module (PLM).}
\label{pixelmoduleoverview}
\end{minipage}%
\begin{minipage}[h]{0.48\linewidth}
\centering
\includegraphics[width=\linewidth]{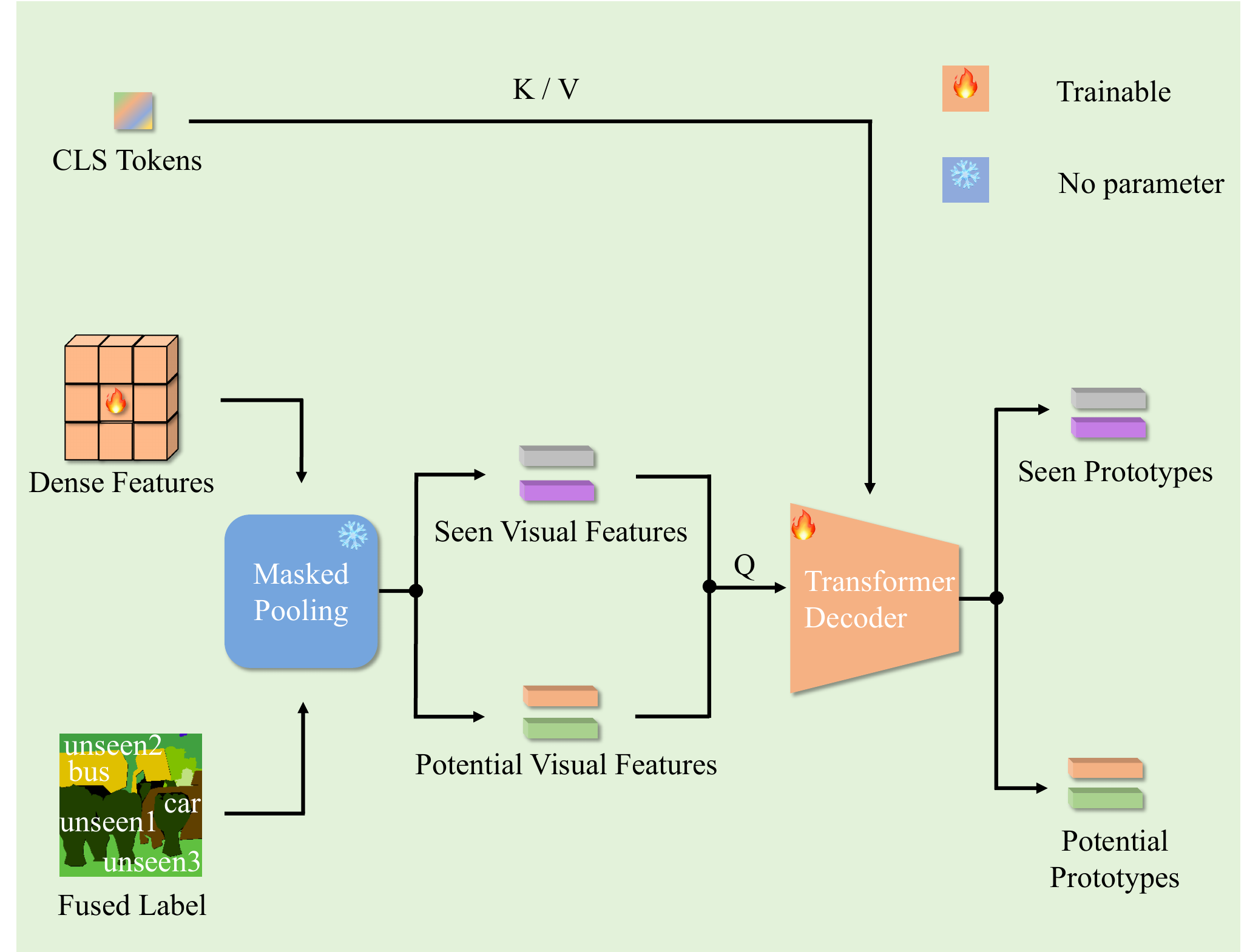}
\subcaption{The structure of Synthesizer. }
\label{VLA}
\end{minipage}
\caption{The structure of PLM where Fig. \ref{VLA} is the details of `Synthesizer' in Fig. \ref{pixelmoduleoverview}.}
\label{pixelmodule}
\vspace{-8pt}
\end{figure*}
\subsection{Pixel Learning Module} \label{PLM}
Though effective, relying only on GLM may not handle the fine-grained segmentation task. Therefore, we propose the Pixel Learning Module (PLM). The structure of PLM is shown in Fig. \ref{pixelmoduleoverview}. PLM consists of a multi-scale K-means algorithm, a mask fusion algorithm to generate the pseudo labels for unannotated areas, and a synthesizer to generate pseudo prototypes for additional supervision to distinguish between seen and unseen categories. 

\noindent \textbf{Pseudo label generation.} Previous research endeavors have found that the features from the models trained by the self-supervised \cite{DINO,dinov2} and unsupervised \cite{SETGO} framework are highly correlated. For the CLIP model, this property has also been studied \cite{PACL}. Motivated by these works, we propose Multi-scale K-means to discover the semantically similar regions for pseudo labels. Precisely, different from GLM, the dense tokens from the CLIP visual encoder are applied as the input. Inspired by the SLIC algorithm \cite{slic}, we first initialize mask seeds in multi-level ways by sliding windows of different sizes to average the dense tokens,

\begin{equation}
  C_d = \sum_{s}^S\sum_{i = 0}^{W / s}\sum_{j = 0}^{H / s}{\frac{D(i+s,j+s)}{(H / s) \cdot (W / s)}},
  \label{densecenter}
\end{equation}
where $H$, $W$ indicates the height and the width of the dense tokens, $S$ indicates the sizes of windows, and $D$ indicates the output dense tokens. Once the seeds $C_d$ are initialized, we apply K-Means based on these seeds only belonging to unannotated areas to produce the pseudo labels. However, only relying on K-Means leads to too small and too many regions. Therefore, we propose a mask fusion algorithm inspired by NMS \cite{focalloss,FCOS} that is widely applied in object detection to remove the redundant bounding boxes. 

Formally, given the $C_d$, its corresponding mask $\mathcal{M}$, and the similarity threshold $\lambda$, we first calculate the cosine similarity $\mathcal{S} \in [-1,1]^{N*N}$ among $C_d$. Then we find the maximum value $\hat{s}_m$ in $\mathcal{S}$ and check whether it is larger than $\lambda$. If lower, we return the $\mathcal{S}$ and treat each seed $\mathcal{S}$ as independent categories with mask $\mathcal{M}$. If larger, the seed $s_m$ is selected to act as the seed for fusing and we find the row index $i$ of $\hat{s}$. Next, all the other values $\hat{S}_i$ larger than $\lambda$ in the $i$th row of $\mathcal{S}$ are selected, and the masks belonging to seeds $\hat{s}_m$ and $\hat{S}_i$ are merged severing as the pseudo label for one unknown category. Note that each seed and mask is used only once, as a result, the rows and columns of $\mathcal{S}$ containing the merged values cannot be used again. Finally, we find new $\hat{s}_m$ and continue to fuse the masks. After obtaining the pseudo labels for unseen categories, we pixel-wisely add the pseudo labels with the seen labels to act as the fused labels $F$ for additional supervision. Different fused masks inherently represent distinct categories without explicit classification like text or auxiliary classifier \cite{pseudoboundingboxlabels,crossmodallabeling}, and these categories might not even appear in the dataset labels, possibly appearing in ``ignored". The pseudo-code of the algorithm is shown in \textbf{\textit{Supplementary materials}}.
             
    
    

\noindent \textbf{Synthesizer. }Despite reliable fused labels being produced, as how many and what categories of the unannotated areas are still unknown, these labels can still not be utilized. To address this issue, we propose a synthesizer based on the fused labels $F$ and the dense features from the image encoder as shown in Fig. \ref{VLA}. Specifically, we first average the dense features based on $F$ to get the centroids of seen and potential categories by averaging the corresponding area,
\begin{equation}
    p_{l} = \frac{\sum_{B,H,W}{{\mathcal{R}}[\mathbb{1}(y_i=l)]}}{\sum_{B,H,W}[\mathbb{1}(y_i=l)]},
\end{equation}
where $B$, $H$, and $W$ respectively indicate the batch size, height, and width of the input images. $y_i$ indicates the label for the $i$th pixel and $l$ indicates the $l$th category in $F$. Then the seen and the potential features, \ie, those extracted from the pseudo label areas, are fed into a transformer decoder as query and the CLS tokens from the CLIP visual encoder as key and value to produce the pseudo prototypes for all the categories. The prototypes are accordingly divided into two parts: seen prototypes $P_{seen}$ and unseen prototypes $P_{pseudo}$. For $P_{seen}$, we use their corresponding text features $T_{seen}$ to supervise,
\begin{equation}
\mathcal{L}_{bce} = \sum_{\substack{p \in P_{seen} \\ t \in T_{seen}}} \log(act(t^Tp))
\label{generateloss}
\end{equation}
where $p \in P_{seen}$ represents the generated seen features, and $t \in T_{seen}$ is the text features from the CLIP text encoder by the name of seen categories. $act$ indicates the sigmoid function. For the $P_{pseudo}$, we apply it as the classifier to further distinguish between the seen and the unseen categories. Formally, the prediction results $x$ are,
\begin{equation}
  x = \text{cat}(\alpha \mathcal{R}^T T_{seen}, \beta \cos(\mathcal{R}, P_{pseudo})),
  \label{logits}
\end{equation}
where $\alpha$ and $\beta$ are two hyper-parameters to control the scale. cat indicates the concatenation along the channel dimension. $\cos(\cdot,\cdot)$ indicates the cosine similarity. Note that as the pseudo labels and the pseudo weights are not exactly precise, cosine similarity avoids outweighing the misclassification and helps distinguish between seen and unseen categories.

\subsection{Training Objective and Inference}
\noindent \textbf{Training Objective. }The training objectives of CLIP-ZSS are:
\begin{equation}
  \mathcal{L} = \mathcal{L}_{global} + \mathcal{L}_{nel}(x, F) + CE(x,F) + \mathcal{L}_{bce},
  \label{inductive loss}
\end{equation}
where $\mathcal{L}_{nel}$ is the same as the NEL loss \cite{zegclip} and $CE$ indicates the cross-entropy.

\noindent \textbf{Inference. } In inference, \textbf{only the image encoder is required} which is the same as the normal semantic segmentation \cite{FCN,segformer,deeplabev3}. Formally, images are fed into the image encoder to obtain the dense features. Then the text features from the CLIP text encoder are applied as the classifier and matrix multiply with the dense features. Finally, the category with max probability is output.

\section{Experiments}
\subsection{Dataset}
To evaluate the effectiveness of our methods, we select three representative benchmarks: PASCAL VOC, COCO-Stuff, and PASCAL Context to conduct our experiments. The split of seen and unseen categories follows the setting of the previous works \cite{zegformer,zegclip,simplebaseline,maskclip}. We introduce the details of these three datasets.

\noindent \textbf{PASCAL VOC} consists of 10,582 images for training and 1,449 images for validation. Note that we convert the `background' category to the `ignored'. For this dataset, there are 15 seen categories and 5 unseen categories.

\noindent \textbf{COCO-Stuff} contains 171 categories totally. As in previous settings, 171 categories are split into 156 seen and 15 unseen categories. Besides, for the training dataset, there are 118,287 images and 5,000 images for testing.

\noindent \textbf{PASCAL Context} includes 4,996 images for training and 5,104 images for testing. For the zero-shot semantic segmentation task, the dataset is split into 49 seen categories and 10 unseen categories.

\subsection{Implementation Details}
The proposed methods are implemented on the open-source toolbox MMsegmentation \cite{mmsegmentation} with Pytorch 2.0.1 \cite{paszke2019pytorch}. The CLIP model applied in our method is based on the ViT-B/16 model and the channel ($C$) of the output text features is 512. All the experiments are conducted on 8 V100 GPUs and the batch size ($B$) is set to 16 for all three datasets. It is worth noticing that our methods can also be applied to the ViT CLIP that can not generate semantically discriminate mask smoothly \cite{clipself,convolutionsdie,fvlm}, and to the dense token from CLIP visual encoder corresponding to the unannotated areas. For all these three datasets, the size of the input images is set as 512 ($H$) $\cdot$ 512 ($W$). The iterations are set to 20k, 40k, and 80k for PASCAL VOC, PASCAL Context, and COCO-Stuff respectively. The optimizer is set to AdamW with the default training schedule in the MMSeg toolbox. In addition, the size of CLS tokens banks is set as 24, the threshold for mask fusion $\lambda$ is 0.8, size of the window in multi-scale K-Means is set as 3 and 7 resulting in 16 and 64 centers for two scales. $\tau$ in global loss is 0.07, and $\gamma$ is 1.5 for COCO-Stuff and 0 for PASCAL VOC and Context. $\alpha$ is set as learnable and $\beta$ is set as 2. The template and the name of unseen categories can be seen in \textbf{\textit{Supplementary materials}}.

To comprehensively evaluate the performance of both seen and unseen categories, we apply the harmonic mean IoU (hIoU) following the previous works \cite{zegclip,zegformer,zs3}. The relationship between mIoU and hIoU is,
\begin{equation}
  hIoU = \frac{2 \cdot mIoU_S \cdot mIoU_U}{mIoU_S + mIoU_U},
  \label{hiou}
\end{equation}
where $mIoU_S$ indicates the mIoU of the seen categories and $mIoU_U$ implies the mIoU of unseen categories. Except for the hIoU, pAcc, and mIoU for seen and unseen categories are also applied in the experiments.

\begin{table*}[t]
\caption{Comparison with state-of-the-art methods where the \textbf{bold} and the \underline{underline} indicates the best and the second-best performance.}
\vspace{-5pt}
\setlength{\tabcolsep}{1pt}
\resizebox{\linewidth}{!}{
\begin{tabular}{cc|cccc|cccc|cccc}
\toprule
\multicolumn{2}{c|}{\multirow{2}{*}{Models}}                & \multicolumn{4}{c|}{PASCAL VOC}                                & \multicolumn{4}{c|}{COCO-Stuff}                                & \multicolumn{4}{c}{PASCAL Context}                                  \\ \cmidrule{3-14} 
\multicolumn{2}{c|}{}                                       & \textbf{pAcc} & \textbf{mIoU(S)} & \textbf{mIoU(U)} & \textbf{hIoU} & \textbf{pAcc} & \textbf{mIoU(S)} & \textbf{mIoU(U)} & \textbf{hIoU} & \textbf{pAcc} & \textbf{mIoU(S)} & \textbf{mIoU(U)} & \textbf{hIoU} \\ \midrule
\multicolumn{2}{c|}{SPNet \cite{spnet}}                                  & -             & 78.0             & 15.6             & 26.1          & -             & 35.2             & 8.7              & 14.0          & -             & -                & -                & -             \\
\multicolumn{2}{c|}{ZS3 \cite{zs3}}                                    & -             & 77.3             & 17.7             & 28.7          & -             & 34.7             & 9.5              & 15.0          & 52.8          & 20.8             & 12.7             & 15.8          \\
\multicolumn{2}{c|}{CaGNet \cite{cagnet}}                                 & 80.7          & 78.4             & 26.6             & 39.7          & 56.6          & 33.5             & 12.2             & 18.2          & -             & 24.1             & 18.5             & 21.2          \\
\multicolumn{2}{c|}{SIGN \cite{sign}}                                   & -             & 75.4             & 28.9             & 41.7          & -             & 32.3             & 15.5             & 20.9          & -             & -                & -                & -             \\
\multicolumn{2}{c|}{Joint \cite{joint}}                                  & -             & 77.7             & 32.5             & 45.9          & -             & -                & -                & -             & -             & 33.0             & 14.9             & 20.5          \\
\multicolumn{2}{c|}{ZegFormer \cite{zegformer}}                              & -             & 86.4             & 63.6             & 73.3          & -             & 36.6             & 33.2             & 34.8          & -             & -                & -                & -             \\
\multicolumn{2}{c|}{zzseg \cite{simplebaseline}}                                  & 90.0          & 83.5             & 72.5             & 77.5          & 60.3          & 39.3             & 36.3             & 37.8          & -             & -                & -                & -             \\
\multicolumn{2}{c|}{ZegCLIP \cite{zegclip}}                                & \textbf{94.6} & \textbf{91.9}    & 77.8             & 84.3          & 62.0          & 40.2             & \textbf{41.4}    & 40.8          & 76.2          & 46.0             & 54.6             & 49.9          \\
\multicolumn{2}{c|}{DeOP \cite{DeOP}}                                   & -             & 88.2             & 74.6             & 80.8          & -             & 38.0             & 38.4             & 38.2          & -             & -                & -                & -             \\ \midrule
\multicolumn{1}{c|}{\multirow{3}{*}{Ours +}} & SegNeXt-B \cite{segnext}    & {\ul 94.0}    & 89.2             & {\ul 82.2}       & {\ul 85.6}    & {\ul 62.8}    & 42.8             & 39.3             & 41.0          & 80.4          & 51.8             & 56.1             & 53.7          \\
\multicolumn{1}{c|}{}                        & Swin-B \cite{swin}      & 93.7          & 88.4             & 81.9             & 85.1          & 62.3          & \textbf{43.8}    & 37.7             & {\ul 40.5}          & {\ul 81.6}    & \textbf{53.5}    & \textbf{65.0}    & \textbf{58.7} \\
\multicolumn{1}{c|}{}                        & Segformer-B4 \cite{segformer} & 93.3          & {\ul 89.6}       & \textbf{83.6}    & \textbf{86.5} & \textbf{63.5} & {\ul 43.3}       & {\ul 41.0}       & \textbf{42.1}          & \textbf{81.7} & {\ul 53.0}       & {\ul 64.9}       & {\ul 58.4}    \\ \bottomrule
\end{tabular}
}
\label{sota}
\vspace{-10pt}
\end{table*}

\subsection{Comparison with State Of The Art}
To demonstrate the effectiveness of our proposed method, we apply three different representative backbones Swin Transformer \cite{swin}, SegNext \cite{segnext} and Segformer \cite{segformer}. Then we compare them with the previous state-of-the-art methods. The results are shown in Table \ref{sota}. From Table \ref{sota}, we can find that our methods can achieve state-of-the-art performance. Specifically, our methods can outperform the existing SOTA methods by a large margin, \ie, 2.2\%, 2.2\%, and 8.5\% in hIoU for PASCAL VOC, COCO-Stuff, and PASCAL Context dataset \wrt ZegCLIP \cite{zegclip} and DeOP \cite{DeOP}. Compared with ZegCLIP, our method achieves better performance in \textbf{hIoU} due to our similar performance in unseen categories and the improved performance for seen categories, indicating that our method achieves better discrimination ability. Meanwhile, all the backbones we use can achieve exceptional improvements. The results above clearly demonstrate the effectiveness of the proposed CLIP-ZSS.

\begin{table}[tbp]
\caption{Ablation experiments of losses.}
\vspace{-5pt}
\setlength{\tabcolsep}{15pt}
\resizebox{\textwidth}{!}{
\begin{tabular}{cccc|cccc}
\toprule
NEL          & CE           & Global   & BCE     & pAcc & mIoU(S) & mIoU(U) & hIoU \\ \midrule
             & $\checkmark$ & $\checkmark$ & $\checkmark$ & 74.5 & 40.4    & 49.7    & 44.6 \\
$\checkmark$ &              & $\checkmark$ & $\checkmark$ & 69.2 & 35.3    & 42.0    & 38.4 \\
$\checkmark$ & $\checkmark$ &              & $\checkmark$ & 51.9 & 28.2    & 8.6    & 13.2 \\
$\checkmark$ & $\checkmark$ & $\checkmark$ &              & \textbf{74.8} & 40.9    & 50.5    & 45.2 \\
$\checkmark$ & $\checkmark$ & $\checkmark$ & $\checkmark$ & \textbf{74.8} & \textbf{41.0}    & \textbf{50.7}    & \textbf{45.4} \\ \bottomrule
\end{tabular}
}
\label{lossablationstudy}
\vspace{-5pt}
\end{table}

\begin{table}[tb]
\begin{minipage}[h]{0.49\linewidth}
\caption{Ablation on designs of GLM.}
\vspace{-10pt}
\setlength{\tabcolsep}{3pt}
\resizebox{\textwidth}{!}{
\begin{tabular}{ccccc}
\toprule
GLM Design & pAcc & mIoU(S) & mIoU(U) & hIoU \\ \midrule
Attention  & \textbf{74.8}    & \textbf{41.0}       & \textbf{50.7}       & \textbf{45.4}    \\
Max        & 71.4    & 38.8       & 32.7       & 35.5    \\
Mean       & 72.7    & 39.6       & 44.7       & 42.0    \\ \bottomrule
\end{tabular}
}
\vspace{5pt}
\label{GLMablationstudy}
\caption{Ablation on designs of PLM.}
\vspace{-10pt}
\setlength{\tabcolsep}{3pt}
\resizebox{\linewidth}{!}{
\begin{tabular}{ccccc}
\toprule
Synthesiser Design  & pAcc & mIoU(S) & mIoU(U) & hIoU \\ \midrule
Transformer Decoder     & 74.8    & \textbf{41.0}       & 50.7       & \textbf{45.4}    \\
MLP               & \textbf{74.9}    & 40.8       & \textbf{50.8}       & 45.2    \\
Original features & 74.6    & 40.6       & 50.6       & 45.1    \\ \bottomrule
\end{tabular}
}
\label{PLMablationstudy}
\vspace{-10pt}
\end{minipage}
\begin{minipage}[h]{0.51\linewidth}
\vspace{10pt}
\centering
\includegraphics[width=1\linewidth]{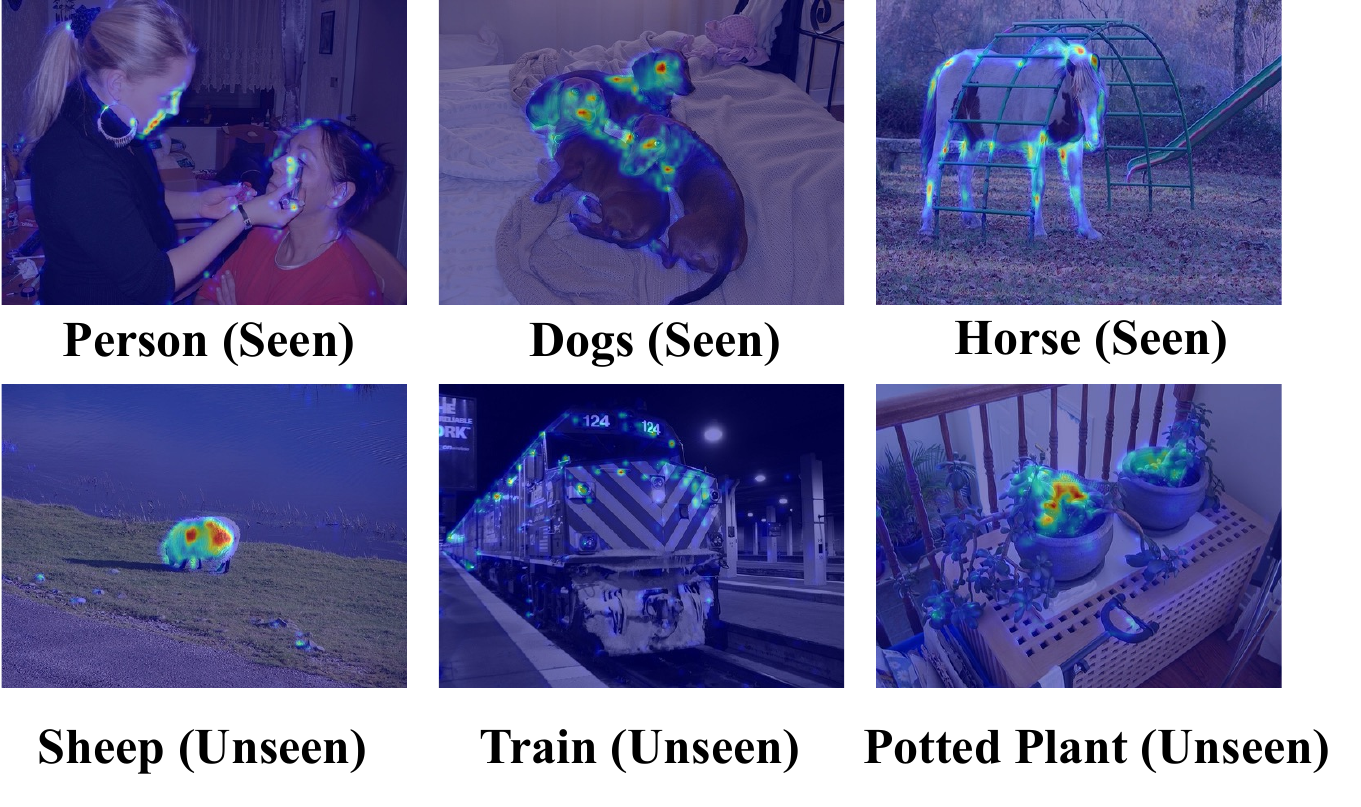}
\captionof{figure}{Attention for seen and unseen.}
\label{heatmap}
\vspace{-10pt}
\end{minipage}
\end{table}


\subsection{Ablation Study}
\label{ablation section}
To evaluate the merits of the proposed methods, we conduct ablation studies. These experiments are conducted in the PASCAL Context with 20k iterations. We use Segformer-B0 as backbones with all the hyperparameters unchanged. 

\noindent \textbf{NEL loss.} We first ablate the NEL loss as shown in the first row of Table \ref{lossablationstudy}. Compared with the model with all the losses, the performance of hIoU drops 0.8\%, \ie, from 45.4\% to 44.9\%. The performance of seen categories drops 0.6\%. The performance for unseen categories drops even lower, \ie, from 50.7\% to 49.7\%. This ablation study implies the effectiveness of NEL loss.

\noindent \textbf{Cross Entropy Loss for PLM.} One of the contributions is providing a novel way to utilize the ignoring areas. As shown in the second row of Table \ref{lossablationstudy}, without the CE, the hIoU drops drastically from 45.4\% to 38.4\%. For seen and unseen categories, the performance is also hurt badly, \ie from 41.0\% to 35.3\%, and 50.7\% to 44.2\% for unseen categories which is also a large drop. This experiment implies the significance of the CE loss.

\noindent \textbf{Global Learning Module.} Another contribution of this work is proposing the GLM module to probe knowledge from the CLIP model. The effectiveness of the GLM is shown in the third row of Table \ref{lossablationstudy}. Note that the $\gamma$ we use in this ablation is 5.0. As depicted, the hIoU drastically drops from 45.4\% to 13.2\% which is a large performance drop. The huge drop is attributed to the weakness in recognition of unseen categories, \ie, unseen mIoU drops from 50.7\% to 8.6\%. This ablation shows the key roles of CLS tokens in the unseen category.

In addition, we further visualize the attention maps of GLM during training as shown in Fig. \ref{heatmap}. In this figure, train, potted plant, and sheep are the unseen categories during training while person, dog, and horse are seen. We can find that even if the models have never seen one category, the model can still find the discriminative parts of the objects for both seen and unseen categories. 

\noindent \textbf{BCE Loss for PLM.} As shown in the fourth row of Table \ref{lossablationstudy}, we do not use any explicit loss function to upgrade the parameters of the synthesizer. The performance drop of 0.2\% in hIoU. Meanwhile, both the performance of seen and unseen categories are hurt, \ie, 0.1\% and 0.2\%, respectively. This result indicates explicit loss function is beneficial.

\noindent \textbf{Different design of GLM.} GLM is proposed to align the dense features with the CLS tokens from the CLIP model. We compare the self-attention-based design with other forms as shown in Table \ref{GLMablationstudy}. In this table, Attention means the original design, max indicates the max pooling in the dense tokens, and mean indicates the average pooling. Our design achieves the highest performance, \ie, 45.4 in hIoU. The performance gain is by a large margin, \ie, 3.4\% higher than the mean and 9.9\% higher than the max. The experiment proves that considering all the features is good to learn from the CLS token.

\noindent \textbf{Ablation study on the threshold of mask fusion} One of the main contributions of this paper is to generate semantically discriminative pseudo-labels for the unannotated areas. In the mask fusion algorithm, the most vital hyperparameter is the threshold to fuse the mask. The ablation study of this hyperparameter is shown in Tab. \ref{thresholdationstudy}. If the threshold is set properly, the performance can reach to expected performance, \eg, when the threshold is 0.9 and 0.8 the performance is 45.6\% and 45.4\%. However, if we fuse most of the mask, \ie, the threshold is set as 0.5 or 0.6, the hIoU drops drastically to 44.7\% and 44.9\%. Especially for the unseen categories, the performance drops by nearly 3.0\%. As a result, the threshold needs to be set properly.
\begin{table}[tb]
\begin{minipage}[h]{0.48\linewidth}
\caption{Ablation on token bank size.}
\vspace{-10pt}
\setlength{\tabcolsep}{5pt}
\resizebox{\linewidth}{!}{
\begin{tabular}{ccccc}
\toprule
Bank size & pAcc & mIoU(S) & mIoU(U) & hIoU \\ \midrule
64  & 74.6    & 40.8       & 49.7       & 44.8    \\
32  & 74.7    & 40.8       & 50.8       & 45.2    \\
24        & 74.8    & \textbf{41.0}       & 50.7       & \textbf{45.4}    \\
16        & 74.9    & 40.7       & \textbf{51.2}       & 45.3    \\
0       & \textbf{75.0}    & 40.8       & 49.0       & 44.5    \\ \bottomrule
\end{tabular}
}

\label{bankablationstudy}
\vspace{-10pt}
\end{minipage}
\begin{minipage}[h]{0.48\linewidth}
\caption{Ablations on fusion thresholds.}
\vspace{-10pt}
\setlength{\tabcolsep}{5pt}
\resizebox{\linewidth}{!}{
\begin{tabular}{ccccc}
\toprule
Threshold & pAcc & mIoU(S) & mIoU(U) & hIoU \\ \midrule
0.9  & 74.6    & 40.9       & \textbf{51.6}       & \textbf{45.6}    \\
0.8        & \textbf{74.8}    & 41.0       & 50.7       & 45.4    \\
0.7       & 74.7    & \textbf{41.2}       & 49.8       & 45.1    \\
0.6       & 74.5    & 41.3       & 49.1       & 44.9    \\
0.5       & 74.4    & 41.2       & 48.8       & 44.7    \\
\bottomrule
\end{tabular}
}
\label{thresholdationstudy}
\vspace{-10pt}
\end{minipage}
\end{table}

\begin{figure}[tb]
\centering
\includegraphics[width=0.8\linewidth]{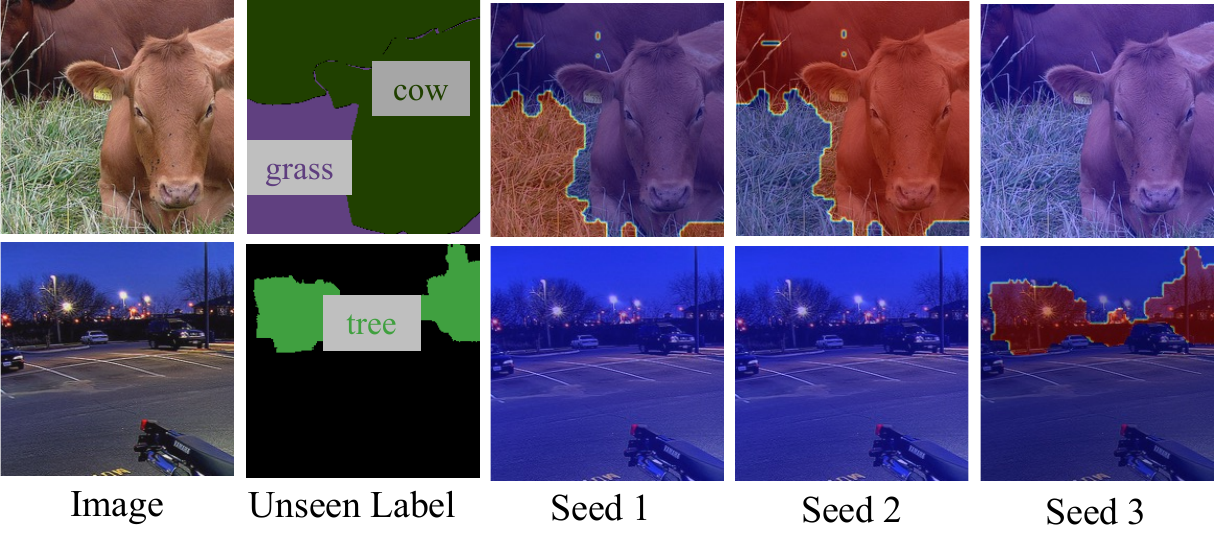}
\vspace{-10pt}
\caption{The pseudo labels generated by PLM.}
\label{kmeansresult}
\vspace{-10pt}
\end{figure}

\noindent \textbf{Different design of synthesizer.} The synthesizer is proposed to utilize both seen and ignored areas. We compare the original design with two other designs: an FC-ReLU-FC network and directly applying the corresponding features as shown in Table \ref{PLMablationstudy}. Compared with the MLP, our design surpasses it by 0.2\% hIoU. For the original features, ours outperforms it by 0.3\%, proving its merits.

\noindent \textbf{Ablation study on different sizes of token bank.} To better probe the knowledge from the CLIP visual encoder, we propose the CLS token bank to increase the number of negative samples. To investigate the effectiveness of the CLS token bank, we conduct experiments on the size of the token bank as shown in Table \ref{bankablationstudy}. The bank size is set from 64 to 0 (no CLS token bank). The hIoU reaches its peak of 45.4\% when the bank size is set to 24 and minimum without the token bank. Besides, the size of the token bank should also be carefully considered, as too many negative pairs may also destroy the performance, \eg, when the size is 64 the hIoU is only 44.8\% which is a great drop from 45.4\%. The experiments above prove the effectiveness of our proposed method.

\begin{table}[tbp]
\centering
\caption{Experiments on the generalization capability. The number after the dataset indicates how many categories are used. }
\vspace{-10pt}
\setlength{\tabcolsep}{10pt}
\resizebox{\textwidth}{!}{
\begin{tabular}{ccc|ccc|c}
\toprule
Method                & BackBone                                          & Training Dataset                & VOC-20        & PC-59         & A-150         & FPS  \\ \midrule
ZS3 \cite{zs3}                 & \multicolumn{1}{c|}{\multirow{4}{*}{ResNet-101 \cite{resnet}}}  & PASCAL VOC \cite{voc}                     & 38.3          & 19.4          & -             & -    \\
LSeg \cite{Lseg}                  & \multicolumn{1}{c|}{}                             & PASCAL VOC \cite{voc}                      & 47.4          & -             & -             & -    \\
OpenSeg \cite{Openseg}              & \multicolumn{1}{c|}{}                             & COCO \cite{coco}                            & 60.0          & 36.9          & 15.3          & -    \\
OpenSeg \cite{Openseg}               & \multicolumn{1}{c|}{}                             & COCO \cite{coco} + Loc. Narr. \cite{narr}              & 63.8          & 40.1          & 17.5          & -    \\ \cmidrule{1-3}
Zzseg-seg             & \multicolumn{1}{c|}{\multirow{2}{*}{ResNet-101c \cite{deeplabev3}}} & \multirow{5}{*}{COCO-Stuff-156 \cite{coco}} & 88.4          & 47.7          & 20.5          & 1.11 \\
DeOP \cite{DeOP}                  & \multicolumn{1}{c|}{}                             &                                 & 91.7          & 48.8          & \textbf{22.9} & 4.37 \\ \cmidrule{1-2} \cmidrule{4-6}
\multirow{3}{*}{Ours} & \multicolumn{1}{c|}{SegNeXt-B \cite{segnext}}                    &                                 & 93.1          & 49.8          & 20.3          & \textbf{30.9} \\
                      & \multicolumn{1}{c|}{Swin-B \cite{swin}}                       &                                 & \textbf{93.6} & 50.0          & 19.9          & 13.3 \\
                      & \multicolumn{1}{c|}{Segformer-B4 \cite{segformer}}                 &                                 & 93.5          & \textbf{50.1} & 20.5          & 21.8 \\ \bottomrule
\end{tabular}
}

\label{generalization}
\vspace{-10pt}
\end{table}

\noindent \textbf{Experiments on Open-Vocabulary Semantic Segmentation. }Moreover, we conduct experiments under the open-vocabulary cross-dataset settings \cite{DeOP,simplebaseline} to evaluate the generalization capability of the proposed methods as shown in Table \ref{generalization}. The metric we use is mIoU. To make a fair comparison, all the models are trained 60k iters and with a batch size of 32 and \textbf{we only use 156 of 171 categories in COCO-Stuff}. As shown in the table, for both the PASCAL Context and the VOC dataset, our methods can still achieve SOTA performance. However, when we test on ADE20k \cite{ade20k}, our performance is lower than DeOP \cite{DeOP}. Meanwhile, without the need for CLIP visual encoder, our method can achieve nearly \textbf{7 times faster} than DeOP while achieving high performance. This phenomenon can be observed on all the backbones we apply, which suggests the high generalization of our methods.

\subsection{Qualitative Results}
\noindent \textbf{The roles of K-Means algorithms.} To further investigate the effectiveness of the pseudo labels, we visualize the results of K-Means and mask fusion algorithms as shown in Fig. \ref{kmeansresult}. The right three images, \ie, Center 1-3, indicate the generated pseudo labels by different cluster centers. As can be seen from the figure, different centers can group different categories. For example, the area grouped by center 1 indicates the grass. For center 2, the cows can be grouped. In center 3 the trees can be found. Note that our method can also discover `novel' categories that are not annotated in the dataset, which can be seen in the supplementary materials. Moreover, though those pseudo labels have high-level semantics, they are not assigned any category information in the dataset. This ablation proves the effectiveness of the multi-scale K-Means and the mask fusion algorithms.

\begin{figure}[tbp]
\centering
\includegraphics[width=0.85\linewidth]{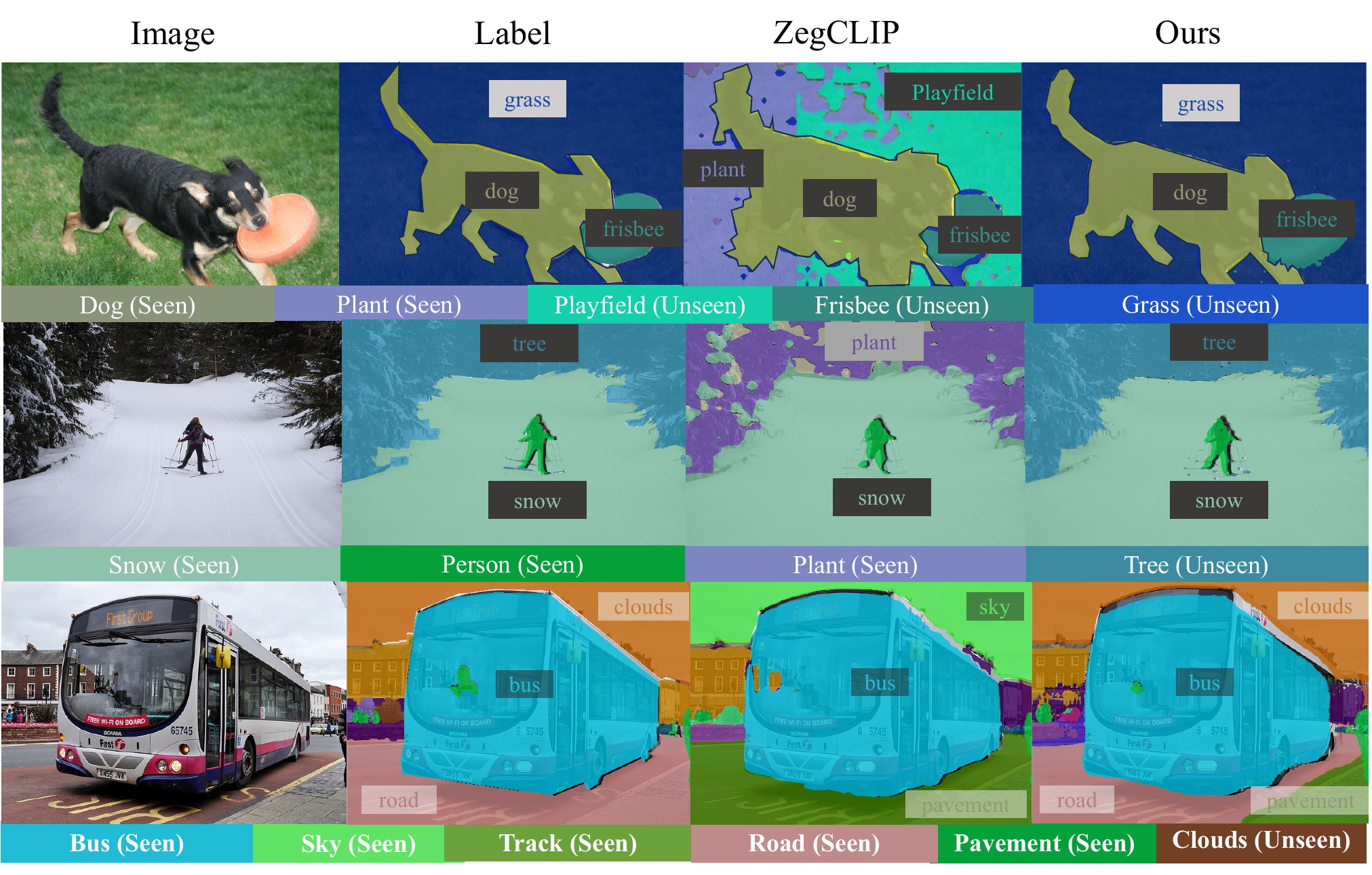}
\caption{The visualization of the prediction.}
\label{visualizationresult}
\vspace{-15pt}
\end{figure}

\noindent \textbf{The visualization of prediction.} We visualize the prediction of our methods as shown in Fig. \ref{visualizationresult}. Compared with SOTA methods, \ie, ZegCLIP \cite{zegclip}, our method can obtain exceptional results on both seen and unseen categories. For example, the `trees' in the second image are classified as very similar plants (seen categories) in ZegCLIP. However, our method can correctly recognize it. More visualizations can be seen in the \textbf{\textit{supplementary materials}}.
\section{Conclusion and Limitations}

In this paper, we proposed CLIP-ZSS, a simple but effective framework that transfers (teaches) the knowledge of CLIP to any image encoder for semantic segmentation without introducing new modules or combinations with VLMs in testing. Specifically, CLIP-ZSS consists of two key modules: Global Learning Module (GLM) and Pixel Learning Module (PLM). GLM is proposed to directly probe knowledge from the CLS token in the CLIP model which is the only token trained in the CLIP visual encoder. PLM can fully take advantage of the input images on both seen and unannotated areas rather than only the seen categories.

However, there are still limitations in our models. The K-Means algorithm in PLM sometimes can not perfectly distinguish between extremely similar categories, \eg, grass, and tree. Moreover, the weight for the ignoring areas is simply produced by a transformer decoder, which may need more careful designs. 
\clearpage  

%
%
\bibliographystyle{splncs04}
\bibliography{egbib}
\end{document}